\pdfoutput=1

\documentclass[11pt]{article}

\usepackage[review]{ACL2023}

\usepackage{times}
\usepackage{latexsym}

\usepackage[T1]{fontenc}

\usepackage[utf8]{inputenc}

\usepackage{microtype}
\usepackage{soul} 
\usepackage{inconsolata}

\usepackage{ragged2e} 
\usepackage{booktabs,makecell, multirow, tabularx}
\usepackage{amsthm,amsmath,amssymb,mathrsfs}
\usepackage{graphicx}
\usepackage{stfloats}

\title{Unlock the Power: Competitive Distillation for Multi-Modal Large Language Models}

\author{
Xinwei Li \\
Southeast University, Nanjing, China\\
  \texttt{seulixinwei@seu.edu.cn} 
  \And
Li Lin \\
Southeast University, Nanjing, China\\
  \texttt{linli321@seu.edu.cn}
  \AND
Shuai Wang \thanks{Corresponding author} \\
Southeast University, Nanjing, China\\
  \texttt{shuaiwang@seu.edu.cn}
  \And
Chen Qian \\
Tsinghua University, Beijing, China\\
  \texttt{qianc62@tsinghua.edu.cn}
}

\begin{document}
\maketitle
\begin{abstract}
Recently, multi-modal content generation has attracted lots of attention from researchers by investigating the utilization of visual instruction tuning based on large language models (LLMs).
To enhance the performance and generalization ability of such LLMs, the practice of distilling knowledge from pretrained multi-modal models (a.k.a. teachers) to more compact multi-modal LLMs (students) has gained considerable interest. 
However, the prevailing paradigm of instruction-tuning in multi-modal LLMs knowledge distillation is resource-intensive and unidirectional, neglecting the potential for mutual feedback between the student and teacher models. 
Thus, we propose an innovative Competitive Multi-modal Distillation framework (\textbf{CoMD}), which captures bidirectional feedback between teacher and student models and continually updates the multi-modal capabilities that the student model has learned.
It comprises two stages: multi-modal pre-training and multi-modal competitive distillation. The first stage pre-trains the student model on a large number of filtered multi-modal datasets. The second stage facilitates a bidirectional knowledge transfer between the student and teacher models. Our experimental analysis of diverse datasets shows that our knowledge transfer method consistently improves the capabilities of the student model. Finally, the 7B-sized student model after four distillations surpassed the current state-of-the-art model LLaVA-13B on the ScienceQA and LLaVA Test dataset, also outperforms other strong baselines in the zero-shot setting.

\end{abstract}

\section{Introduction}
\begin{figure}[htbp]
\centering 
\includegraphics[width=1.0\linewidth]{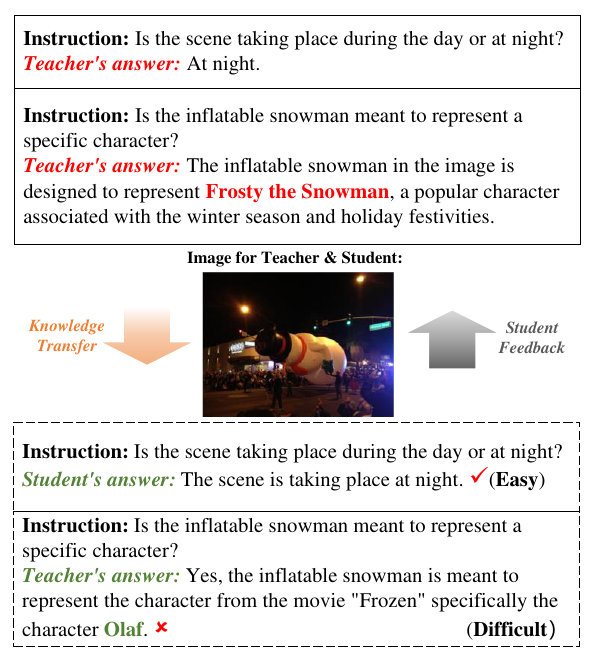}
\caption{A comparison of our approach with previous methods.} 
\label{Fig.1}
\end{figure}
In recent years, there has been a steady increase in the parameter size and pre-training data scale of LLMs, leading to the continuous surpassing of the upper limit of natural language understanding capabilities \cite{chen2020big,radford2019language,chowdhery2022palm,zhang2022opt,raffel2020exploring}. 
To further enhance LLMs' capabilities, 
researchers focus on the development of open-source multi-modal LLMs, the prevailing paradigm, called instruction-tuning, to distill multi-modal knowledge by aligning the responses of the more compact LLMs (student models) with those of the larger size LLMs (teacher models) in response to a set of instructions, such as BLIP-2 \cite{blip2}, LLaVA \cite{llava}, and MiniGPT-4 \cite{minigpt4}.\par 
However, the instructions are usually generated through GPT-4 or based on a manually constructed dataset, constructing multi-modal instructions using these methods can be expensive or labor-intensive \cite{selfinstruct}. Furthermore, the instruction tuning-based knowledge transfer method is unidirectional. As shown by the orange arrow in Figure \ref{Fig.1}, The instructions constructed from the teacher model's answers enable the student model to learn the ability to distinguish “whether the scene in the current image the day or at night". However, the student model still struggles to answer difficult questions, such as "identifying the specific cartoon figure corresponding to the snowman in the image". However, current model distillation methods do not take such student's feedback into account, as shown by the gray arrow in Figure \ref{Fig.1}, the feedback refers to the identification of difficult instruction where the student model is performing poorly. Incorporating the feedback ensures that the teacher model can provide customized training focused on addressing these difficult examples, thereby enhancing the capabilities of the student model.\par
\begin{figure*}[htbp]
\centering 
\includegraphics[width=1.0\textwidth]{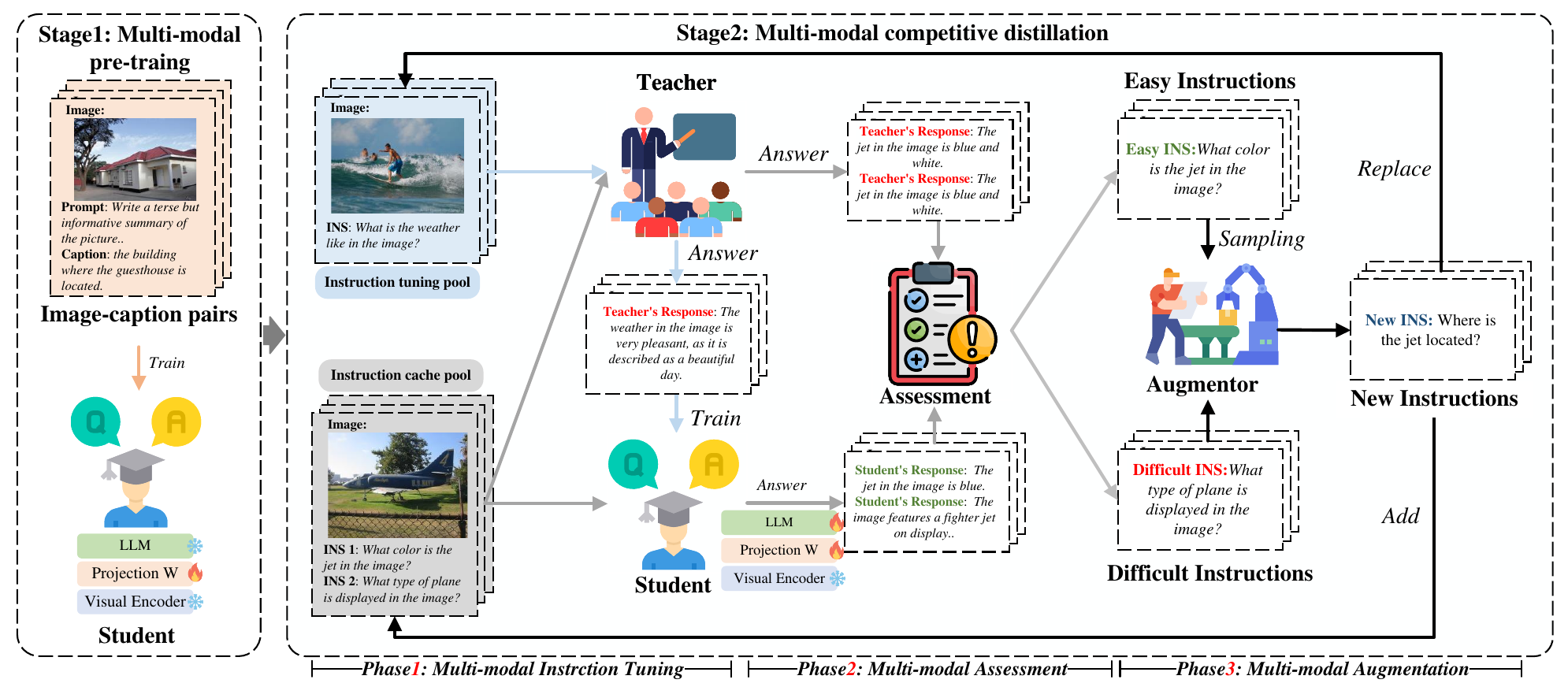}
\caption{The overview of our multi-modal competitive distillation framework. From left to right, there are two stages, and the second stage consists of three phases in an iteration: 1) Multi-modal instruction tuning; 2) Multi-modal assessment; 3) Multi-modal augmentation.} 
\label{Fig.2}
\end{figure*}

To address the two main challenges above, we propose a novel framework for multi-modal large model knowledge distillation, our framework is depicted in Figure \ref{Fig.2}, consisting of two stages: stage 1:  Multi-modal pre-training, which aims to train a projection layer to align multi-modal features. Stage 2: Multi-modal competitive distillation, consists of three phases in an iteration: 1) Multi-modal instruction tuning, which aligns student responses with multi-modal instructions given by the teacher; 2) Multi-modal-assessment, which identifies difficult multi-modal instructions; and 3) Multi-modal-Augmentation, which generates more new instructions and combine them with original images to build a new multi-modal instruction dataset to train the student model. Essentially, the multi-modal competitive distillation stage establishes a bidirectional feedback loop that effectively enhances the multi-modal capabilities of the student model.\par
To evaluate the effectiveness of our method, we employ our \textbf{Co}mpetitive \textbf{M}ulti-modal \textbf{D}istillation framework to transfer the knowledge from LLaVA-13B to our 7B-sized student model  (\textbf{CoMD}), which shares the same model architecture as the teacher model. Our dataset was initialized based on llava-80K (contains only images and corresponding instructions without answers). We conduct three iterations of reasoning and result in 504K multi-modal data that our model is trained on.
The experimental results that our knowledge transfer method consistently improves the capabilities of student models, and superior performance surpassing multi-modal large language model as LLaVA \cite{llava}. Our main contributions are as follows:
\begin{itemize}
\item[$\bullet$] Our work is the first attempt to adopt the idea of competitive distillation to open-source multi-modal large language models.
\item[$\bullet$] Our proposed framework demonstrates impressive efficiency and effectiveness. Initializing the dataset without any human annotations, our model outperforms the current SOTA model on the reasoning task and outperforms models with larger parameter sizes in the zero-shot setting.
\item[$\bullet$] The versatility of our framework allows for a wide range of applications, and it can be easily adapted to fit a variety of other open-source multi-modal large language models.
\end{itemize}

\section{Related Work}
\subsection{Multi-modal Instruction Tuning}
Instructions-tuning aims to train LLM by utilizing datasets with diverse NLP tasks. This effective method has been successfully applied to well-known LLMs like InstructGPT \cite{instructgpt} and FLAN-T5 \cite{flant5}, significantly improving their performance and generalization capabilities. Building on this success, instruction-tuning extended to the visual domain recently. MiniGPT4 \cite{minigpt4} utilizes ChatGPT to enhance the detailed description of image captions and generate high-quality instruction data. LLaVA \cite{llava} generates multi-modal instruction data by prompting plain text GPT-4 \cite{gpt4} along with bounding boxes of objects and image captions. LLaMA-Adapter \cite{llamaadapter,llamaadapterv2} aligns text and image features using the COCO dataset and leverages only text data for instruction tuning. mPLUG-owl \cite{mplug} pre-trains the model with over 1000M image-text pairs and constructs a 400M hybrid dataset comprising plain text and multi-modal instruction fine-tuning data. InstructBLIP \cite{instructblip} converts 13 visual language tasks into multi-modal instruction tuning data format for instruction tuning. 
Such work usually constructs instruction datasets through closed-source large models or manual efforts, which is costly and labor-intensive.
Therefore, it is crucial to prioritize the quality of instructions over their quantity as this can enhance the capability of the multi-modal model and reduce the cost of constructing instruction data.
\subsection{Knowledge distillation}
Knowledge distillation (KD) aims to transfer knowledge from a teacher model to a student model. Currently, knowledge distillation can be classified into two categories: black-box distillation and white-box distillation \cite{dismodel}. In black-box KD, the student model only has access to the teacher's predictions, while white-box KD allows the student model to utilize the weights of the teacher model \cite{dismodel}. Typically, the prevailing distillation method for large language models is black-box distillation, which can be divided into three subcategories: In context learning (ICL) distillation \cite{incontext,incontext2}, Chain-of-Thought (CoT) distillation \cite{cotprompt,selfsonsistency,cotreasoners}, and Instruction Following (IF) distillation \cite{trainhuman,Instructpix2pix,lion}. \cite{ildis} introduced the ICL distillation method, which transfers contextual few-shot learning and language modeling abilities from the teacher model to the student model. In contrast, CoT distillation takes a different approach, MT-COT \cite{mtcot} aims to enhance the reasoning performance of the student model by utilizing the CoT generated by the teacher model.
Step-by-Step distillation \cite{stepbystep} employs chain-of-thought arguments generated by LLM as additional guidance for training student models within a multi-task framework. IF distillation tuning the model using a series of complex NLP tasks presented as instructions. LaMini-LM \cite{lamamini} model utilizes chatgpt as its teacher model, and generates new instruction by prompting chatgpt, resulting in a comprehensive dataset comprising 2.58 million instructions, covering a diverse array of topics.
Although these methods successfully distill the knowledge of teacher models into the student models, they still strictly follow the unidirectional knowledge transfer without considering what can teachers learn from students about how to teach effectively.
Furthermore, they are not applicable to multimodal domains. Therefore, a competitive distillation method is essential for capturing multi-modal feedback from both students and teachers and continuously updating the knowledge learned by the student model.
\begin{figure*}[htbp]
\centering 
\includegraphics[width=0.8\textwidth]{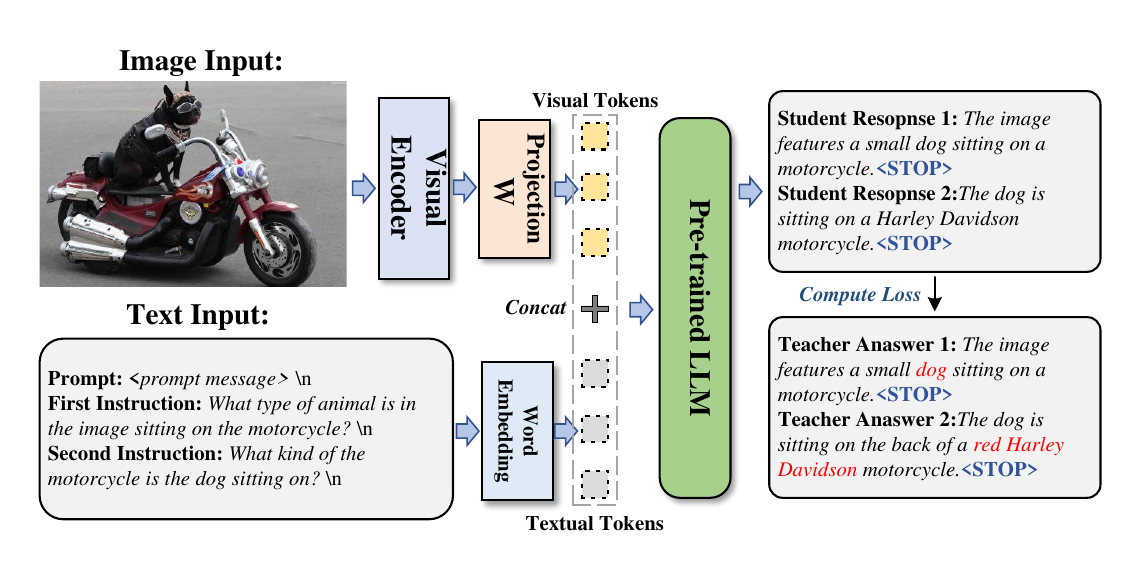}
\caption{The architecture of our model (CoMD) and multi-turn dialogue instruction data training method, only two turns of dialogue as an example illustrated here.} 
\label{Fig.3}
\end{figure*}

\section{Methodology}
The objective of our work is to utilize the outputs of current open source multi-modal LLM (teacher model $\mathcal{T}$) to iteratively distill a high-performing multi-modal student model $\mathcal{S}$. We first introduce the architecture and training method of $\mathcal{S}$. Then we illustrated our multi-modal distillation method, which consists of two stages: the multi-modal pre-training stage and the multi-modal competitive distillation stage. In the first stage, we freeze the visual encoder and LLM, and pre-train the feature alignment layer (projection matrix $\mathbf{W}$) using a large number of image-text pairs. The purpose of this stage is to train a visual tokenizer for the frozen LLM. The second stage is a three-phase cycle: 1) the multi-modal instruction tuning phase is designed to transfer knowledge from a continuously updated dataset from the teacher model to the student model; 2) the multi-modal assessment phase aims to judge the difficulty of instruction data and obtain feedback information from the student model; and 3) the multi-modal augmentation phase is responsible to generate novel instructions to consistently present new challenges to the student model.
During the training of the student model $\mathcal{S}$, the student will master difficult multi-modal instructions and subsequently convert them into simpler instructions.

\subsection{Model architecture and training method}
\label{3.1}
We design a unified multi-modal model architecture to accept both textual and image inputs. Figure \ref{Fig.2} illustrates the architecture of our student model $\mathcal{S}$. For the textual input, \textbf{CoMD} is initialized using Vicuna-7B-1.1 \cite{vicuna}, which has been fine-tuned using supervised data based on LLaMa-7B \cite{llama}. For the image input, we utilize the pre-trained CLIP visual encoder ViT-L/14 \cite{clip} to extract the visual features. The architecture of the teacher model is similar to that of our student model, with the LLM initialized with Vicuna-13B-1.1 \cite{vicuna}, and the same visual encoder being employed.\par
The training data for our student model can be divided into two categories: single-turn dialogue instruction data and multi-turn dialogue instruction data. We unify the formats of the two types of training data to facilitate model training in both stages without the need to modify the model architecture. Specifically, the training data $D$ containing $N$ multi-modal instruction-tuning data $X_i$ can be denoted as:
\begin{equation} 
\begin{aligned}
&D=\left\{\left(X_i\right)\right\}_{i \in[1, N]} \\
&X_i=\left(V_i, C_i\right) \\
&C_i=\left\{\left(Q_t, A_t\right)\right\}_{t \in[1, M]}
\label{eq1}
\end{aligned} 
\end{equation} 

Here $C_i$ represents the $i$-th multi-modal instruction data, and $M$ represents the total number dialogues in the $i$-th turn. $A_i$ represents the corresponding answer. We organize $C_i$ at the $t$-th turn as a unified format sentence $C_{i}^{t}$:
\begin{equation} 
\begin{aligned} 
C_{i}^{t}=\begin{cases}\left[P,V_{i},Q_{1}\right] \text{  or }\left[P,Q_{1},V_{i}\right], &t=1 \\ Q_{t},&t>1\end{cases}
\end{aligned} 
\end{equation} \par
Figure \ref{Fig.3} illustrates a specific example of training using two turns of instruction data. $V_i$ represents the image input, $P$ represents the prompt of the student model $\mathcal{S}$, $Q_i$ corresponds to the instruction of the $i$-th turn, and $A_i$ represents the answer. The student model $\mathcal{S}$ is also trained to predict the answers and determine where to stop, so we add the <STOP> token to indicate the end of an instruction or answer as shown in Figure \ref{Fig.3}. Consequently, the loss of our model is computed using only the response from the student model $\mathcal{S}$, and the <STOP> token.\par
Specifically, we merge each $\{(V_i, C_i)\}$ into an image-text pair sequence, then we compute the probability of generating target answers $A^S_t$ by:

\begin{small}
\begin{equation} 
\begin{aligned} 
p\left(A^S_t \mid V_{i}, Q_{t}\right)=\prod_{t=1}^M p_{\boldsymbol{\theta}}\left(x_t \mid V_{t}, Q_{t,<m}, A^S_{t,<m}\right)
\end{aligned}
\end{equation} 
\end{small}

Here $\theta$ represents the trainable parameter of the student model. $Q_{t,<m}, A^S_{t,<m}$ refer to the questions and answer tokens generated by the model in all previous turns before the current prediction tokens of the student model. Additionally, we explicitly include $V_i$ to emphasize that all responses are image-based. For better readability, we have omitted prompt $P$ and all previous occurrences of <STOP>, although they are also used as conditional information for the responses of the student model.

\subsection{Multi-modal pre-training stage}
The first multi-modal pre-training stage aims to train the feature alignment layer using a large number of filtered image-text pairs. To accomplish this, we filter the combined dataset from Conceptual Caption 3M \cite{cc12m}, SBU \cite{sbu}, and LAION \cite{laion}. Specifically, we employ Spacy to extract noun phrases from each caption in the combined dataset and calculate the frequency of each phrase. Noun phrases with frequencies less than 3 are excluded since they typically represent rare concepts and attributes that are already covered by other pairs. Pairs containing these excluded noun phrases are sequentially added to the candidate pool, starting with the noun phrase with the lowest remaining frequency. If a noun phrase occurs more than 100 times, we randomly select a subset of 100 pairs that contain that noun phrase. By applying this filtering method, the combined dataset yields approximately 885K image-text pairs.\par
As shown in Figure \ref{Fig.3}, The filtered dataset is then used to train the alignment matrix $\mathbf{W}$ of the student model $\mathcal{S}$, and keep the weights of the visual encoder and pre-trained LLM frozen. This stage can be interpreted as the process of training a visual tokenizer for the student model $\mathcal{S}$.

\subsection{Multi-modal competitive distillation stage}
The second multi-modal competitive knowledge distillation stage consists of three phases: 1) Multi-modal instruction tuning phase, which aligns students' responses with teachers' responses; 2) Multi-modal Assessment, which identifies difficult instructions; and 3) Multi-modal-Augmentation, which generates instructions to increase the challenges faced by student models. Figure \ref{Fig.3} illustrates the establishment of four roles and two data pools in our framework. We initialize the teacher model $\mathcal{T}$, Assessmentor $\mathcal{R}$, and Augmentor $\mathcal{G}$ using the same multi-modal open-source large model, i.e., LLaVA-13B \cite {llava}. We prompt LLaVA-13B \cite{llava} so that it plays different roles in three phases.\par
Our data pools are built on LLaVA-80K \cite{llava}, which consists of 80,000 multi-turn dialogue data. This dataset can also be represented as Formula \ref{eq1}. where $V_i$ represents the picture corresponding to the multi-turns dialogue $C_i$. Here, $N$ represents the total number of multi-turns dialogues included in the dataset, which is 80,000. $Q_t$ represents the question based on the picture content in each turn of dialogue, while $A_t$ represents the answer generated by GPT-4 to the question. $M$ represents the number of turns in the multi-turn dialogue. To initialize our multi-modal instruction tuning pool, we first convert LLaVA-80K to single-turn dialogue data, and then we remove the answers in the dialogue. This results in a single-turn multi-modal question dataset, denoted as $D_{T}$:

\begin{equation} 
D_{T}=\{(V_k,X_k)\}_{k \in\left[1,K\right]}
\end{equation} 

The initialization of our instruction cache pool is the same as the instruction tuning pool. It is used to store all instructions for evaluating the multi-modal reasoning performance of both the student model and the teacher model.
\subsubsection{Multi-modal instruction tuning phase}
In this phase, we prompt LLaVA-13B \cite{llava} as teacher model $\mathcal{T}$, and generate corresponding answer $A^{T}_{k}=\mathcal{T}\left(V_k, X_k\right)$ for each multi-modal question in instruction tuning pool. Then we convert all single-turn dialogues $\left(V_k, X_k, A^{T}_{k}\right)$ into multi-turn dialogue forms as in formula \ref{eq1}. We use the training method in subsection \ref{3.1} to instruction tuning our student model $\mathcal{S}$.

\subsubsection{Multi-modal assessment phase}
Figure \ref{Fig.3} illustrates the multi-modal assessment phase, which begins with the instruction cache pool, denoted as $D_C$. While the instruction tuning pool and the instruction cache pool have the same initial state, their purposes differ. The instruction tuning pool is refreshed by replacing its current instructions with newly generated ones, whereas the instruction cache pool is refreshed by merging all newly generated instructions to store all instructions. 
Based on the data in the cache pool, we prompt LLaVA-13B \cite{llava} as an assessment to assess the difficulty of multi-modal instruction based on the responses from the teacher model $\mathcal{T}$ and the student model $\mathcal{S}$. To accomplish this, we input each multi-modal instruction from the cache pool into both the $\mathcal{T}$ and $\mathcal{S}$, then we prompt each to generate an answer. Subsequently, we prompt the assessment to score the answers provided by the $\mathcal{T}$ and $\mathcal{S}$:
\begin{equation} 
\begin{aligned}
R_k^S=\mathcal{A}\left(\mathcal{S}\left(V_k,X_k\right) \mid (V_k,X_k,\mathcal{T}\left(V_k,X_k\right)\right))\\
R_k^T=\mathcal{A}\left(\mathcal{T}\left(V_k,X_k\right) \mid (V_k,X_k,\mathcal{S}\left(V_k,X_k\right)\right))
\label{formula5}
\end{aligned}
\end{equation} \par
The construction of this prompt is inspired by the prompt template proposed by \cite{vicuna}. The prompt necessitates that the LLM comprehensively evaluate the two answers based on their usefulness, relevance, accuracy, level of detail, and output in a specified format. 
To mitigate any positional bias of the LLM referee \cite{llmeva}, we repeat the process twice by exchanging the positions of the teacher's response and the student's response. The final score is then calculated as the average of the two runs. Once the scores are obtained, we employ formula \ref{formula6} to calculate the degree of difficulty. 
\begin{equation} 
\begin{aligned}
S_k=\frac{a b s\left(R_k^S-R_k^T\right)+1}{\max \left(R_k^S, R_k^T\right)}
\label{formula6}
\end{aligned}
\end{equation} \par
This formula first calculates the absolute value of the difference between the scores of student and teacher, then normalizes it to the range of 0 to 1, so that the impact of the score difference on different score levels is consistent. Finally, we obtain a score that can reflect the difficulty of the problem. The higher the score, the more difficult the instruction. We add 1 to the difference to avoid the situation when the numerator equals to 0. For example, for instruction $Q_m$ and $Q_n$, $R_m^S$=1, $R_m^T$=1, and $R_n^S$=9, $R_n^T$=9. Obviously, $Q_n$ is much more difficult than $Q_m$, but both scores are 0. Based on $S_k$, We set a threshold $\tau=0.33$ to classify instructions into two categories: difficult instructions, denoted by $S_k \geq \tau$, and easy instructions, denoted by $S_k < \tau$.

\subsubsection{Multi-modal augmentation phase}
After assessing the difficulty of the multi-modal instructions, the objective of the generation phase is to produce new instructions that differ in content but are similar in difficulty to the original pictures. This process is conducted by prompting the open-source multi-modal large model, referred to as the augmentor $\mathcal{G}$. The identified difficult instructions are inputted, and prompt $\mathcal{G}$ generates a new instruction based on each difficult instruction and its corresponding picture. The newly generated instructions must align with the task type of the original instruction and possess a significant difficulty coefficient. In order to alleviate catastrophic forgetting of the model and enhance the diversity of instructions tuning pool. We sample the set of identified easy instructions such that the number of difficult and easy instructions is equal. These instructions are then used to prompt the generator in the same manner to generate new instructions.\par
To ensure instruction diversity, each newly generated instruction is considered valid only if its ROUGE-L score with all other instructions of the corresponding image is below 0.7. Finally, as described in Figure \ref{Fig.3}, the original instructions in the tuning pool are replaced with the new instructions, while simultaneously enriching the cache pool by incorporating the newly generated instructions.
\section{Experiments}
\subsection{Experimental Settings}
In our experiment, we comprehensively evaluated the multi-modal student model after three iterations of distillation. We considered two tasks: fine-tuning on downstream datasets and zero-shot inference, which include various capabilities such as complex reasoning, scene understanding, and scientific question answering.
\subsubsection{Datasets}
\textbf{ScienceQA} \cite{scienceqa} is a large-scale multi-modal dataset utilized for scientific question-answering, encompassing a wide range of domains, including three subjects, 26 themes, 127 categories, and 379 skills. ScienceQA is composed of plain text and text-image examples, divided into three segments: training, validation, and testing, containing 12,726, 4,241, and 4,241 examples respectively.\par
\textbf{SEED-Bench} \cite{seedbench} 
includes 19K multiple-choice questions, covering 12 evaluation dimensions across image and video modalities. We chose the image modality (SEED-Bench IMG) to evaluate our model under a zero-shot setting, which includes 9 dimensions and 14K multiple-choice questions. \par
\textbf{LLaVA Test Set} \cite{llava} comprises 90 multi-modal questions, covering three categories: conversation, complex reasoning, and detail description. Primarily, the LLaVA Test Set evaluates the performance of the model in multi-modal conversations.

\begin{table*}
\centering
\caption{Comparison of the performance (accuracy \%) of CoMD on the scienceQA dataset with other powerful baseline models and SOTA models. Question classes: NAT = natural science, SOC= social science, LAN =  language science, TXT = text context, IMG = image context, NO = no context, G1-6 = grades 1-6, G7-12 = grades 7-12.}
\resizebox{\textwidth}{!}{
\begin{tabular}{l|ccc|ccc|cc|c}
\toprule[1pt]
\multicolumn{1}{c|}{\multirow{2}{*}{\textbf{Method}}} & \multicolumn{3}{c|}{\textbf{Subject}}                                              & \multicolumn{3}{c|}{\textbf{Context Modality}}                                     & \multicolumn{2}{c|}{\textbf{Grade}}                    & \multirow{2}{*}{\textbf{Average}} \\
\multicolumn{1}{c|}{}                                 & NAT                       & SOC                       & LAN                        & TXT                       & IMG                       & NO                         & G1-6                      & G7-12                      &                                   \\ \midrule[1pt]
Human                                                 & 90.23                     & 84.97                     & 87.48                      & 89.60                     & 87.50                     & 88.10                      & 91.59                     & 82.42                      & 88.40                           \\
GPT-3.5 \cite{chatgpt}                                               & 74.64                     & 69.74                     & 76.00                      & 74.44                     & 67.28                     & 77.42                      & 76.80                     &68.89                      & 73.97                             \\
GPT-3.5 w/ CoT \cite{chatgpt}                                        & 75.44                     & 70.87                     & 78.09                      & 74.68                     & 67.43                     & 79.93                      & 78.23                     & 69.68                      & 75.17                             \\
GPT-4 \cite{gpt4}                                                 & \multicolumn{1}{l}{84.06} & \multicolumn{1}{l}{73.45} & \multicolumn{1}{l|}{87.36} & \multicolumn{1}{l}{81.87} & \multicolumn{1}{l}{70.75} & \multicolumn{1}{l|}{90.73} & \multicolumn{1}{l}{84.69} & \multicolumn{1}{l|}{79.10} & 82.69                             \\
LLaMA-Adapter \cite{llamaadapter}                                         & 84.37                     & 88.30                     & 84.36                      & 83.72                     & 80.32                     & 86.90                      & 85.83                     & 84.05                      & 85.19                             \\
MM-CoT$_{Base}$ \cite{mmcot}                                           & 87.52                     & 77.17                     & 85.82                      & 87.88                     & 82.90                     & 86.83                      & 84.65                     & 85.37                      & 84.91                             \\
MM-CoT$_{Large}$ \cite{mmcot}                                           &  \textbf{95.91}               & 82.00                     &  \textbf{90.82}                &  95.26               & \underline{88.80}                     &  \textbf{92.89}                & \underline{92.44}                     & 90.31                      &  \underline{91.68}                       \\
LLaVA \cite{llava}                                                 & 90.36                     &  \underline{95.95}                     & 88.00                      & 89.49                     & 88.00                     & 90.66                      & 90.93                     & \underline{90.90}                      & 90.92                             \\ \midrule[1pt]
\textbf{CoMD}                                                 &  \underline{91.83}                     & \textbf{95.95}            &  \underline{88.91}                     & \textbf{90.91}                     & \textbf{89.94}            & 91.08                      & \textbf{92.47}            & \textbf{90.97}             & \textbf{91.94}                    \\ \bottomrule[1pt]
\end{tabular}
}
\label{tablescience}
\label{tab1}
\end{table*}
\subsubsection{Baselines}
For ScienceQA dataset, we select powerful VQA models: MM-CoT Base \&Large \cite{mmcot}, the current SOTA model LLaVA-13B \cite{llava} (also our teacher model) and the Open AI GPT model (LLaMA-Adapter \cite{llamaadapter}, GPT3.5 \cite{chatgpt}, GPT-4 \cite{gpt4}) as our baseline method. For the text-only baseline, we use the image caption to prompt the model. \par
For SEED-Bench dataset and LLaVA Test Set, we choose the mainstream 7B size multi-modal LLMs, including Otter \cite{otter}, OpenFlamingo \cite{Openflamingo}, MultiModal-GPT \cite{mmgpt}, mPLUG-Owl \cite{mplug}, LLaMA-Adapter V2 \cite{llamaadapterv2}, InstructBLIP \cite{instructblip}, GVT \cite{GVT}, VisualGLM \cite{visual6B}, MiniGPT-4 \cite{minigpt4}, Ziya-Visual \cite{ziyavisual} and our teacher model LLaVA-13B \cite{llava}as our baseline models.
\subsubsection{Implementation Details}
Our multi-modal competitive distillation framework underwent a total of four iterations in the second stage. Based on student feedback and the bidirectional competition between the teacher model and student model, our instruction tuning pool sequentially increased by 220K (initialized based on LLaVA 220K single-turn multi-modal instructions, without answers), 84K, 90K, and 110K instruction-tuning data respectively, which resulted in our model being trained sequentially for 4 Epochs. The instruction-tuning pool contains a total of 504K multi-modal instruction data, and the pre-training data set contains a total of 885K multi-modal dialogue data.\par
We adopt AdamW as the optimizer, with the batch size and warmup ratio set to 16 and 0.03, respectively. For the first and second stages, we set the learning rate to 2e-3 and 2e-5, respectively. In our multi-modal knowledge transfer framework, the temperature of the Teacher, Assessment, and Augmentor all are 0.5. For the ScienceQA dataset, we trained on the training dataset for 6 epochs and set the learning rate to 2e-5, keeping the remaining hyperparameters unchanged, and in the testing phase, for ScienceQA, SEED-Bench, the temperatures are 0.5, 0.1 respectively.\par
All our experiments were conducted on 6 V100 (32G), and we used DeepSpeed \cite{deepspeed}and Xformer \cite{xformer}to optimize GPU memory usage.
\begin{table*}
\caption{Comparison of the performance (accuracy \%) of CoMD on the SEED-Bench dataset with other 7B multi-modal LLMs and 13B LLaVA. Question classes: SUG =  Scene Understanding, IIY =  Instance Identity, IAS = Instance Attributes,  ILN = Instance Location, ICG = Instance Counting, SRS = Spatial Relations, IIN = Instance Interaction, VRG = Visual Reasoning, TRN = Text Recognition.}
\resizebox{\textwidth}{!}{
\centering
\begin{tabular}{ll|ccccccccc|c}
\toprule[1pt]
\multicolumn{1}{c|}{\textbf{Model}}       & \multicolumn{1}{c|}{\textbf{Language Model}} & \textbf{SUG}   & \textbf{IIY}   & \textbf{IAS}   & \textbf{ILN}   & \textbf{ICG}   & \textbf{SRS}   & \textbf{IIN}   & \textbf{VRG}   & \textbf{TRN}   & \textbf{Average} \\ \midrule[1pt]
\multicolumn{1}{l|}{Otter \cite{otter}}               & LLaMA-7B                                     & 44.90          & 38.56          & 32.24          & 30.88          & 26.28          & 31.81          & 31.96          & 51.36          & 31.76          & 35.16            \\
\multicolumn{1}{l|}{OpenFlamingo \cite{Openflamingo}}        & LLaMA-7B                                     & 43.86          & 38.12          & 31.28          & 30.06          & 27.30          & 30.59          & 29.90          & 50.15          & 20.00          & 34.51            \\
\multicolumn{1}{l|}{MultiModal-GPT \cite{mmgpt}}      & LLaMA-7B                                     & 43.64          & 37.85          & 31.45          & 30.78          & 27.34          & 30.14          & 29.90          & 51.36          & 18.82          & 34.54            \\
\multicolumn{1}{l|}{mPLUG-Owl \cite{mplug}}           & LLaMA-7B                                     & 49.68          & 45.33          & 32.52          & 36.71          & 27.26          & 32.72          & 44.33          & 54.68          & 18.82          & 37.88            \\
\multicolumn{1}{l|}{LLaMA-Adapter V2 \cite{llamaadapterv2}}    & LLaMA-7B                                     & 45.22          & 38.50          & 29.30          & 33.03          & 29.67          & 35.46          & 39.18          & 51.96          & 24.71          & 35.19            \\
\multicolumn{1}{l|}{InstructBLIP Vicuna \cite{instructblip}} & Vicuna-7B                                    & 60.20          & \textbf{58.93} & \textbf{65.63} & \textbf{43.56} & \textbf{57.05} & \textbf{40.33} & \textbf{52.58}          & 47.73          & \underline{43.53}          & \textbf{58.76}   \\
\multicolumn{1}{l|}{GVT \cite{GVT}}                 & Vicuna-7B                                    & 41.74          & 35.50          & 31.79          & 29.45          & \underline{36.17}          & 31.96          & 31.96          & 51.06          & 27.06          & 35.49            \\
\multicolumn{1}{l|}{LLaVA \cite{llava}}               & Vicuna-13B                                   & \textbf{63.43}          & 49.10          & 49.04          & \underline{43.04}          & 30.93       & \underline{38.35}          & 45.36 & \underline{61.32}          & 38.82          & 48.43            \\ \midrule[1pt]
\multicolumn{1}{l|}{\textbf{CoMD}}               & Vicuna-7B                                    & \underline{63.10} & \underline{51.50}          & \underline{53.80}          & 42.23          & 34.36          & 38.20          & \underline{51.54}          & \textbf{64.35} & \textbf{47.05} & \underline{50.90}            \\ \bottomrule[1pt]
\end{tabular}
}
\end{table*}

\subsection{Experimental Results}

\subsubsection{ScienceQA}
In Table \ref{tab1}, we first compare \textbf{CoMD} with the strong baseline methods and the current SOTA models. From this table, we observe that the current LLMs, such as GPT3.5 (COT) \cite{chatgpt}, GPT4 \cite{gpt4}, still underperform compared to humans in few-shot or zero-shot settings, indicating that ScienceQA still presents a significant challenge for these models. In contrast, existing supervised methods yield better results. \par
Notably, MM-CoT Large \cite{mmcot} achieves previous state-of-the-art results, with an average accuracy of 91.68\%. LLaVA-13B \cite{llava}, serves as our teacher model and adopts a model architecture similar to ours, which is more closely aligned with our work. The results suggest that LLaVA \cite{llava}remains competitive compared to MM-CoT Large \cite{mmcot}, particularly in the SOC category. Our model achieves better feature alignment by pre-training on a richer dataset (885K compared to LLaVA's 556K). Importantly, based on our multi-modal competitive distillation framework, \textbf{CoMD} is trained on a larger and higher-quality instruction dataset, allowing it to surpass LLaVA's performance in nearly all categories with fewer parameters (7B compared to LLaVA's 13B). Furthermore, \textbf{CoMD} outperforms the current SOTA method, MM-CoT Large \cite{mmcot}, in the SOC, IMG, G1-6, G7-12 categories, as well as in the final average accuracy. This makes it the first model of 7B size to surpass MM-CoT Large \cite{mmcot}. By prompting the teacher model in our method, the teacher model plays various roles, generating more high-quality instruction data and transferring more knowledge to the student model. The results validate the significant effectiveness of our multi-modal competitive distillation framework.
\subsubsection{SEED-Bench}

We evaluated the multi-modal reasoning performance of \textbf{CoMD} in the zero-shot setting on the SEED-Bench dataset \cite{seedbench}. We selected the mainstream 7B-size model and our teacher model LLaVA-13B as our baseline models. The results demonstrate that our proposed model, \textbf{CoMD}, exhibits competitive performance, achieving the highest accuracy in visual reasoning and text recognition. It surpasses the current SOTA model, InstructBLIP \cite{instructblip}, by 16.62\% and 3.52\%, respectively, as well as the teacher model LLaVA-13B \cite{llava} by 3.03\% and 8.82\%, respectively. This underlines the superior visual reasoning and text recognition capabilities of CoMD, which is attributed to our multi-modal knowledge distillation framework. Through continual iterative distillation, CoMD is trained with more instruction data, enhancing the model's understanding of different scenarios and text instructions. However, its performance in instance location and spatial relations is slightly inferior to InstructBLIP Vicuna \cite{instructblip} and LLaVA \cite{llava}. This may be due to LLaVA's larger parameter size, which is advantageous for fine-grained spatial position recognition, and InstructBLIP Vicuna's larger multi-modal instruction dataset (16M), which enables the model to learn more visual knowledge.\par
Our model ranks second in average accuracy among all current 7B-size models, next to the InstructBLIP \cite{instructblip}. The superiority of InstructBLIP \cite{instructblip}is mainly due to its tuning data, which includes 16M multi-modal samples (30 times more than ours), covering a wide range of multi-modal tasks, including OCR and visual reasoning QA data. Our work does not require the construction of a large instruction dataset through manual labor or other closed-source large models, hence our main contribution is orthogonal to that of InstructBLIP \cite{instructblip}. Our multi-modal distillation framework is applicable to other open-source large models, allowing the performance to be continually improved at a minimal cost.\par
Finally, we found that multi-modal LLMs still perform poorly on fine-grained visual reasoning tasks, such as Instance Counting, Spatial Relations, Instance Interaction, and Text Recognition. This suggests that fine-grained visual question-answering tasks still pose a significant challenge to multi-modal large models. However, our CoMD shows improvement in fine-grained tasks compared to the teacher model LLaVA-13B \cite{llava}, with increases in performance of 3.43\%, 6.18\%, and 8.23\% respectively. This indicates that our method can help enhance the performance of models on fine-grained visual tasks.
\subsubsection{LLaVA Test Set}

\begin{table}
\caption{Comparison of the results (Score rated by GPT-4) of CoMD on the LLaVA Test Set with other powerful baselines. Question classes: Con: conversation category. CR: complex reasoning category. DD: detail description category}
\resizebox{\linewidth}{!}{
\begin{tabular}{l|ccc|c}
\toprule[1pt]
\multicolumn{1}{c|}{\textbf{Model}} & \textbf{Con}  & \textbf{CR}   & \textbf{DD}   & \textbf{AVG}  \\ \midrule[1pt]
VisualGLM \cite{visual6B}          & 65.8          & 80.6          & 64.5          & 70.3          \\
MiniGPT-4 \cite{minigpt4}         & 65.3          & 75.6          & 66.3          & 69.1          \\
mPLUG-owl \cite{mplug}          & 69.0          & 84.1          & 59.0          & 70.8          \\
Ziya-Visual \cite{ziyavisual}      & 82.3          & 90.2          & 71.2          & 81.3          \\
InstructBLIP \cite{instructblip}      & 82.2          & 90.2          & 68.4          & 80.7          \\
LLaVA \cite{llava}             & \underline{83.1}          & \textbf{96.5} & \underline{75.3}          & \underline{85.1}          \\ \hline
\textbf{CoMD}                               & \textbf{86.4} & \underline{93.0}          & \textbf{77.5} & \textbf{85.7} \\ \bottomrule[1pt]
\end{tabular}
}

\end{table}
As displayed in Table 3, the results of our proposed model, \textbf{CoMD}, are compared with other leading baseline models on the LLaVA Test Set across three question categories: conversation (Con), complex reasoning (CR), and detail description (DD). The scores were rated by GPT4.\par
In the conversation category, \textbf{CoMD} outperforms all other models with a score of 86.4. For the complex reasoning category, the highest score of 96.5 is achieved by LLaVA (Liu et al. 2023), with \textbf{CoMD} scoring slightly lower at 93.0. When it comes to the detail description category, \textbf{CoMD} again leads with a score of 77.5.  This may be attributed to the fact that \textbf{CoMD}, in the iterative distillation process, is trained with more complex instructions (including detailed description tasks) as well as simpler instructions (conversation tasks). As a result, it is able to provide more detailed descriptions of images and generate richer dialogue content.
Taking the average scores into account, \textbf{CoMD} demonstrates superior overall performance with a score of 85.7, slightly surpassing LLaVA's average score of 85.1 and other baseline models.\par

In summary, our proposed model, \textbf{CoMD}, exhibits robust performance across all categories, especially in conversation and detail description categories, indicating its effectiveness and versatility in handling different types of questions in the context of natural language processing.
\subsubsection{Ablation Results}
\begin{table}
\caption{Ablation study of the threshold $\tau$ for CoMD.}

\resizebox{\linewidth}{!}{
\begin{tabular}{l|ccc}
\toprule[1pt]
\multicolumn{1}{c|}{\textbf{Threshold $\tau$}}
                               & \begin{tabular}[c]{@{}c@{}}\textbf{Science}\\ \textbf{QA}\end{tabular} & \begin{tabular}[c]{@{}c@{}}\textbf{SEED-}\\ \textbf{Bench}\end{tabular} & \begin{tabular}[c]{@{}c@{}}\textbf{LLaVA} \\ \textbf{Test Set}\end{tabular} \\ \midrule[1pt]
0 (\textbf{w/o} easy. Inst.) & 91.08                                                & 48.21                                                 & 85.10                                                      \\
0.33 (\textbf{Ours})                           & \textbf{91.94}                                                & \textbf{50.90}                                                  & \textbf{85.70}                                                      \\
0.67                           & 91.32                                                & \underline{49.37}                                                 & 84.80                                                      \\
1 (\textbf{w/o} diff. Inst.)      & 91.06                                                & 48.79                                                 & \underline{85.30}                                                      \\ \bottomrule[1pt] 

\end{tabular}
}
\label{tab4}
\end{table}
\textbf{The parameter $\tau$ that differentiates between difficult and easy instructions.} As shown in Table \ref{tab4}, We conducted a systematic investigation of $\tau$ ranging from 0.0 to 1.0, observing its impact on average performance across three datasets. $\tau$ = 0 implies that all newly generated instructions are considered difficult, thus excluding any simple instructions. Conversely, $\tau$ =1 implies that all newly generated instructions are considered simple, thereby excluding any difficult instructions. The experimental results demonstrate that a lack of diversity in difficult and simple instructions can lead to decreased model performance. Remarkably, our model exhibits optimal performance when $\tau$ = 0.33. This suggests that our parameter settings effectively discriminate between difficult and easy instructions.\par
\begin{table}
\caption{Ablation study of multi-modal pre-training stage for CoMD.}
\resizebox{\linewidth}{!}{
\begin{tabular}{c|ccc}
\toprule[1pt]
\textbf{Method}
                               & \begin{tabular}[c]{@{}c@{}}\textbf{Science}\\ \textbf{QA}\end{tabular} & \begin{tabular}[c]{@{}c@{}}\textbf{SEED-}\\ \textbf{Bench}\end{tabular} & \begin{tabular}[c]{@{}c@{}}\textbf{LLaVA} \\ \textbf{Test Set}\end{tabular} \\ \midrule[1pt]
\textbf{w/o} Pre-training & 86.43              & 42.31               & 78.88          \\ \bottomrule[1pt]
\end{tabular}

}
\label{table5}
\end{table}

\textbf{Multi-modal pretraining stage.} We skip the pre-training phase and utilize all the generated instructions datasets to train the base model, maintaining consistent other parameters. As shown in Table \ref{table5}, it is evident that the model's accuracy on the three datasets has significantly declined (-5.51\%, -8.59\%, -6.82\%). This underscores the significance of our pre-training stage in preserving a substantial amount of pre-training knowledge while concurrently aligning multi-modal features.

\subsubsection{Effect of number of iterations}
\begin{figure}[htbp]
\centering 
\includegraphics[width=1.0\linewidth]{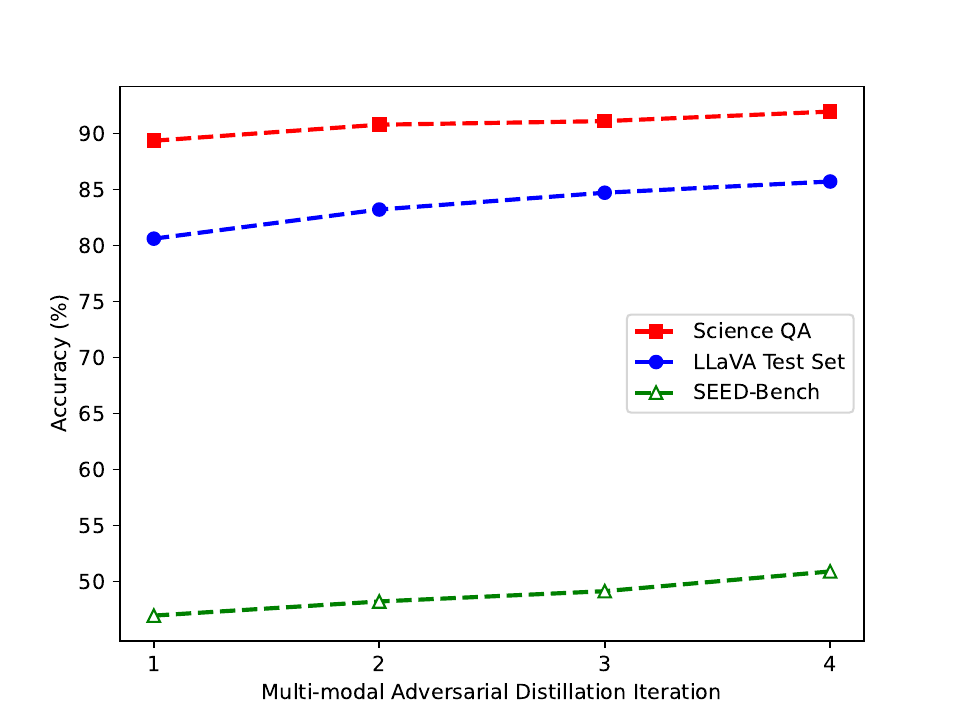}
\caption{Performance of CoMD on ScienceQA, SEED-Bench, and LLaVA Test Set through the Distillation iterations.} 
\label{Fig.5}
\end{figure}

Figure \ref{Fig.5} illustrates the performance of \textbf{CoMD} on the ScienceQA, SEED-Bench, and LLaVA test sets over four distillation iterations. The results indicate a consistent enhancement of the student model's performance as the number of iterations escalates, with the most substantial improvement occurring in the initial iteration. This result underscores the efficacy of our multi-modal three-stage competitive distillation framework.

\section{Case Study}
\subsection{Qualitative Comparison}
\begin{figure*}[htbp]
\centering 
\includegraphics[width=0.9\textwidth]{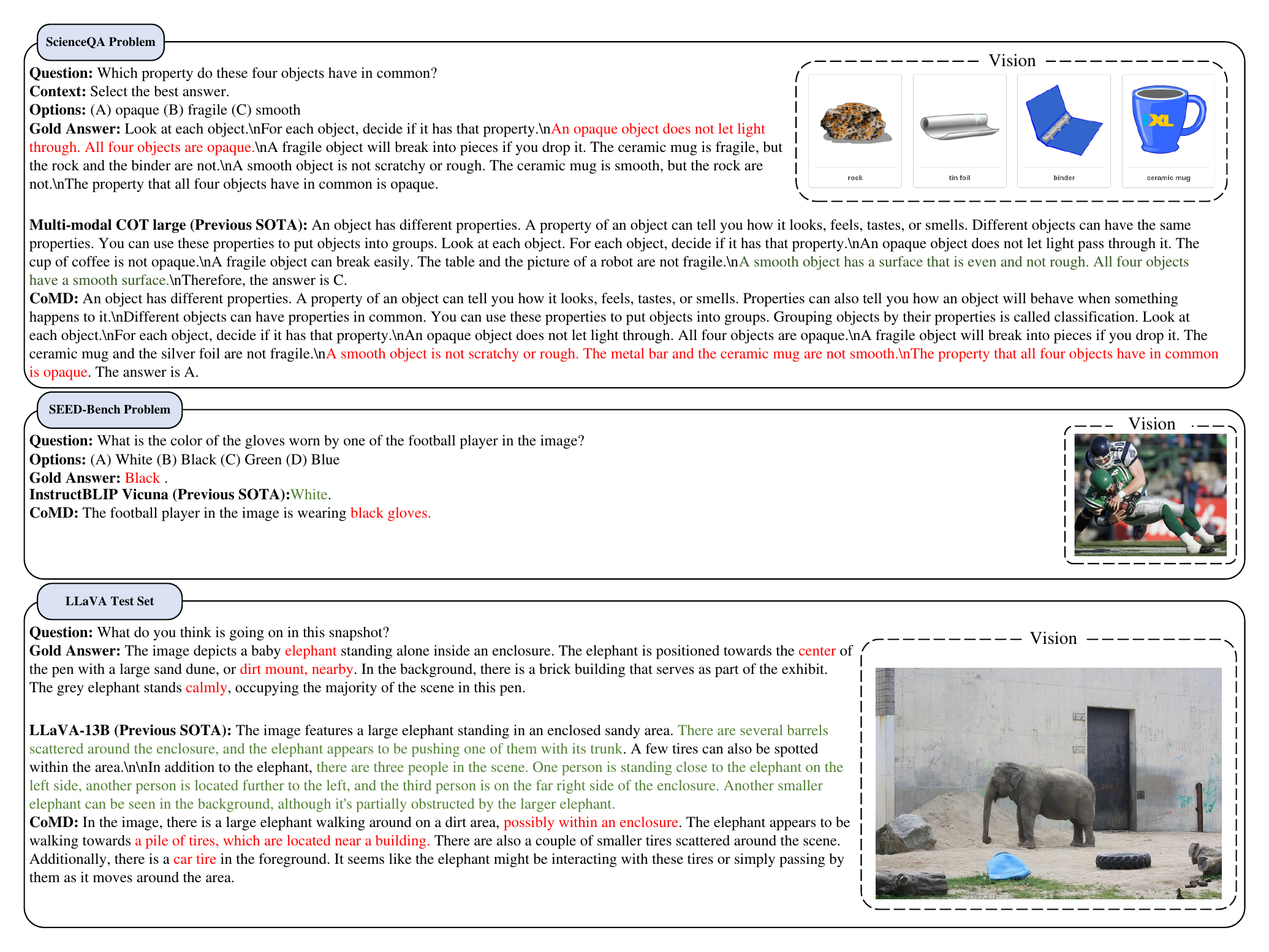}
\caption{Case analysis of multi-modal questions. we demonstrate the performance of our CoMD model in comparison with previous SOTA models on selected samples across three different datasets.} 
\label{Fig.6}
\end{figure*}

 To better understand the multi-modal capabilities of our \textbf{CoMD} model, we selected representative examples from three different datasets and compared the responses of the \textbf{CoMD} model with those of the previous SOTA models within each dataset. The comparison results are shown in the Figure \ref{Fig.6}.\par
We found that the \textbf{CoMD} model is more adept at analyzing fine-grained elements within an image. For example, in the first sample, the task is to analyze four different objects in the image (rock, tin foil, binder, ceramic mug) and infer their common characteristics. While the Multi-modal COT large model analyzed the four objects, it failed to correctly deduce their shared characteristics. In contrast, the \textbf{CoMD} correctly inferred that "An opaque object does not let light through. All four objects are opaque."\par
In the second example, the task involves perceiving complex motion states, and then determining the color of a sportsman's gloves. The InstructBLIP model incorrectly identified the color of one sportsman's wristband (white) as the answer. However, the \textbf{CoMD} correctly identified and discerned the color of the gloves (black) in the image, demonstrating superior comprehension of complex images.

The third example requires a detailed description of the image content. We observed that while the LLaVA model described some of the main elements in the image (elephant, sandy area), its answer is hallucinatory, including a non-existent person as a key element in its description. On the other hand, the \textbf{CoMD} avoided this issue, accurately described the main elements (elephant, dirt area), and also identified the "tire" element that was missing in the answer.
\subsection{Error Analysis}
\begin{figure*}[htbp]
\centering 
\includegraphics[width=0.9\textwidth]{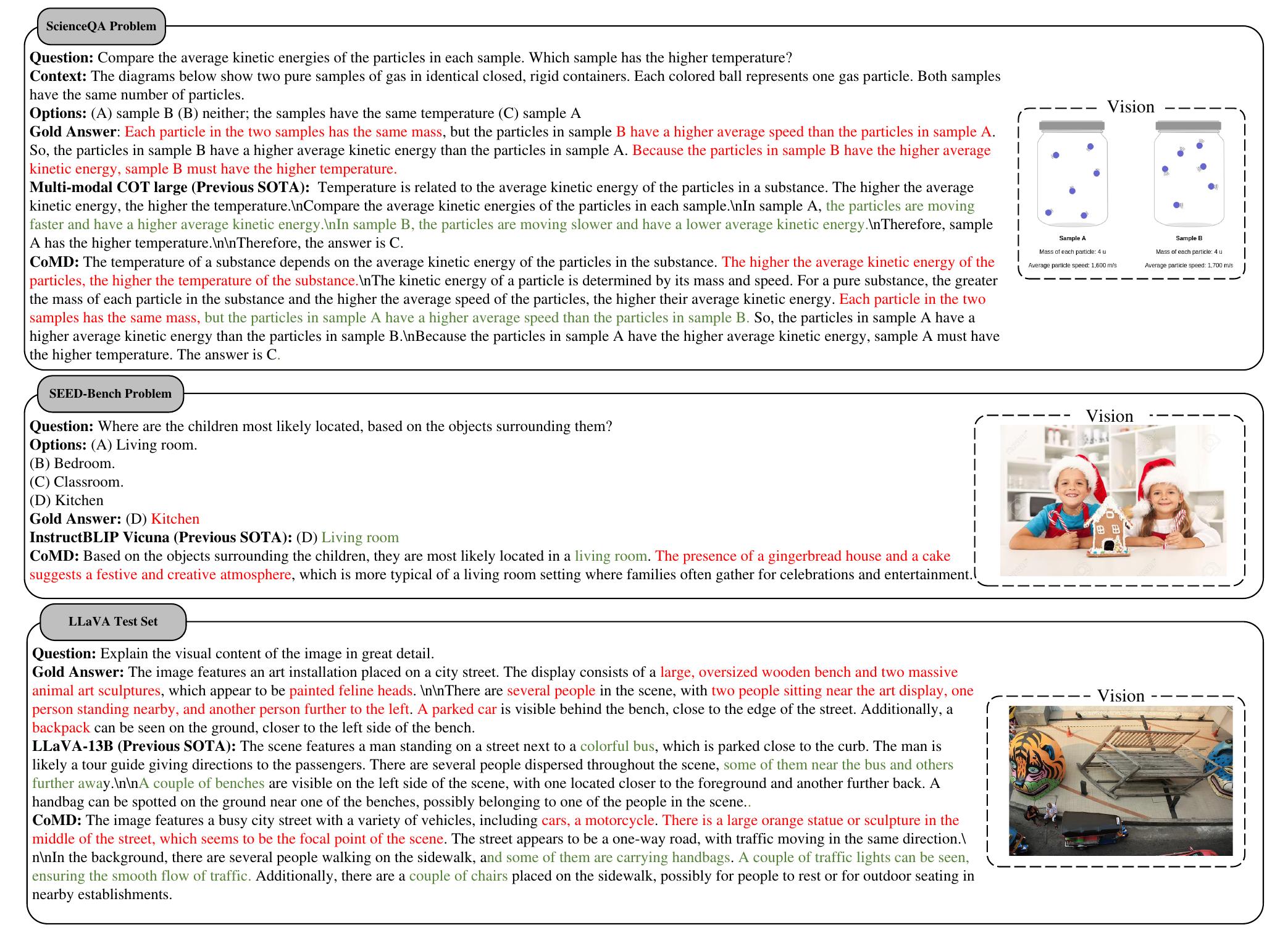}
\caption{Examples of different types of errors under three datasets are presented, with the correct answers marked in red and incorrect responses in green. The three categories of examples, from top to bottom are: counting numbers, event reasoning, and detail description.} 
\label{Fig.7}
\end{figure*}

To further understand the behavior of the \textbf{CoMD} and facilitate future studies, we present some errors that both the \textbf{CoMD} and previous SOTA models tend to make, as illustrated in Figure \ref{Fig.7}.\par
In the first example, although neither multi-modal COT large nor \textbf{CoMD} identified the average velocity information of the particles, \textbf{CoMD} provided a more detailed inference path. In the second example, \textbf{CoMD} accurately identified the gingerbread house in the image, but due to the "celebrations and entertainment" information associated with this element, \textbf{CoMD} mistakenly inferred that the scene in the image took place in a living room. In the third example, facing a complex image description task, while \textbf{CoMD} lacked descriptive details about the state and behavior of the people in the image, it correctly described the main elements in the image, including "cars, motorcycle, truck, large orange statue."

\section{Conclusion}
This paper proposes a novel framework for multi-modal large model knowledge distillation, addressing the challenge of expensive and labor-intensive construction of multi-modal instructions and the unidirectional nature of instruction tuning-based knowledge transfer. Our method introduces a bidirectional feedback loop through multi-modal competitive distillation, effectively enhancing the student model's capabilities. The effectiveness of our framework was validated through experiments, demonstrating consistent enhancement of student models' capabilities and superior performance compared to existing multi-modal LLMs such as LLaVA. Our work is the first to apply competitive distillation to open-source multi-modal LLMs, demonstrating impressive efficiency and efficacy, and versatility for a wide range of applications. With the ability to initialize the dataset without any human annotations, our model outperforms the current state-of-the-art models on the reasoning task and larger parameter size models in the zero-shot setting. This framework can be adapted to fit a variety of other open-source multi-modal LLMs, paving the way for further advancements in this field.
\bibliography{anthology,custom}

\begin{thebibliography}{48}
\expandafter\ifx\csname natexlab\endcsname\relax\def\natexlab#1{#1}\fi

\bibitem[{Awadalla et~al.(2023)Awadalla, Gao, Gardner, Hessel, Hanafy, Zhu,
  Marathe, Bitton, Gadre, Sagawa et~al.}]{Openflamingo}
Anas Awadalla, Irena Gao, Josh Gardner, Jack Hessel, Yusuf Hanafy, Wanrong Zhu,
  Kalyani Marathe, Yonatan Bitton, Samir Gadre, Shiori Sagawa, et~al. 2023.
\newblock Openflamingo: An open-source framework for training large
  autoregressive vision-language models.
\newblock \emph{arXiv preprint arXiv:2308.01390}.

\bibitem[{Brooks et~al.(2023)Brooks, Holynski, and Efros}]{Instructpix2pix}
Tim Brooks, Aleksander Holynski, and Alexei~A Efros. 2023.
\newblock Instructpix2pix: Learning to follow image editing instructions.
\newblock In \emph{Proceedings of the IEEE/CVF Conference on Computer Vision
  and Pattern Recognition}, pages 18392--18402.

\bibitem[{Changpinyo et~al.(2021)Changpinyo, Sharma, Ding, and Soricut}]{cc12m}
Soravit Changpinyo, Piyush Sharma, Nan Ding, and Radu Soricut. 2021.
\newblock Conceptual 12m: Pushing web-scale image-text pre-training to
  recognize long-tail visual concepts.
\newblock In \emph{Proceedings of the IEEE/CVF Conference on Computer Vision
  and Pattern Recognition}, pages 3558--3568.

\bibitem[{Chen et~al.(2020)Chen, Kornblith, Swersky, Norouzi, and
  Hinton}]{chen2020big}
Ting Chen, Simon Kornblith, Kevin Swersky, Mohammad Norouzi, and Geoffrey~E
  Hinton. 2020.
\newblock Big self-supervised models are strong semi-supervised learners.
\newblock \emph{Advances in neural information processing systems (NeurIPS)},
  33:22243--22255.

\bibitem[{Chiang et~al.(2023)Chiang, Li, Lin, Sheng, Wu, Zhang, Zheng, Zhuang,
  Zhuang, Gonzalez et~al.}]{vicuna}
Wei-Lin Chiang, Zhuohan Li, Zi~Lin, Ying Sheng, Zhanghao Wu, Hao Zhang, Lianmin
  Zheng, Siyuan Zhuang, Yonghao Zhuang, Joseph~E Gonzalez, et~al. 2023.
\newblock Vicuna: An open-source chatbot impressing gpt-4 with 90\%* chatgpt
  quality.
\newblock \emph{See https://vicuna. lmsys. org (accessed 14 April 2023)}.

\bibitem[{Chowdhery et~al.(2022)Chowdhery, Narang, Devlin, Bosma, Mishra,
  Roberts, Barham, Chung, Sutton, Gehrmann et~al.}]{chowdhery2022palm}
Aakanksha Chowdhery, Sharan Narang, Jacob Devlin, Maarten Bosma, Gaurav Mishra,
  Adam Roberts, Paul Barham, Hyung~Won Chung, Charles Sutton, Sebastian
  Gehrmann, et~al. 2022.
\newblock Palm: Scaling language modeling with pathways.
\newblock \emph{arXiv preprint arXiv:2204.02311}.

\bibitem[{Chung et~al.(2022)Chung, Hou, Longpre, Zoph, Tay, Fedus, Li, Wang,
  Dehghani, Brahma et~al.}]{flant5}
Hyung~Won Chung, Le~Hou, Shayne Longpre, Barret Zoph, Yi~Tay, William Fedus,
  Yunxuan Li, Xuezhi Wang, Mostafa Dehghani, Siddhartha Brahma, et~al. 2022.
\newblock Scaling instruction-finetuned language models.
\newblock \emph{arXiv preprint arXiv:2210.11416}.

\bibitem[{Dai et~al.(2023)Dai, Li, Li, Tiong, Zhao, Wang, Li, Fung, and
  Hoi}]{instructblip}
Wenliang Dai, Junnan Li, Dongxu Li, Anthony Meng~Huat Tiong, Junqi Zhao,
  Weisheng Wang, Boyang Li, Pascale Fung, and Steven Hoi. 2023.
\newblock \href {http://arxiv.org/abs/2305.06500} {Instructblip: Towards
  general-purpose vision-language models with instruction tuning}.

\bibitem[{Dong et~al.(2022)Dong, Li, Dai, Zheng, Wu, Chang, Sun, Xu, and
  Sui}]{incontext}
Qingxiu Dong, Lei Li, Damai Dai, Ce~Zheng, Zhiyong Wu, Baobao Chang, Xu~Sun,
  Jingjing Xu, and Zhifang Sui. 2022.
\newblock A survey for in-context learning.
\newblock \emph{arXiv preprint arXiv:2301.00234}.

\bibitem[{Gao et~al.(2023)Gao, Han, Zhang, Lin, Geng, Zhou, Zhang, Lu, He, Yue
  et~al.}]{llamaadapterv2}
Peng Gao, Jiaming Han, Renrui Zhang, Ziyi Lin, Shijie Geng, Aojun Zhou, Wei
  Zhang, Pan Lu, Conghui He, Xiangyu Yue, et~al. 2023.
\newblock Llama-adapter v2: Parameter-efficient visual instruction model.
\newblock \emph{arXiv preprint arXiv:2304.15010}.

\bibitem[{Gong et~al.(2023)Gong, Lyu, Zhang, Wang, Zheng, Zhao, Liu, Zhang,
  Luo, and Chen}]{mmgpt}
Tao Gong, Chengqi Lyu, Shilong Zhang, Yudong Wang, Miao Zheng, Qian Zhao,
  Kuikun Liu, Wenwei Zhang, Ping Luo, and Kai Chen. 2023.
\newblock Multimodal-gpt: A vision and language model for dialogue with humans.
\newblock \emph{arXiv preprint arXiv:2305.04790}.

\bibitem[{Hsieh et~al.(2023)Hsieh, Li, Yeh, Nakhost, Fujii, Ratner, Krishna,
  Lee, and Pfister}]{stepbystep}
Cheng-Yu Hsieh, Chun-Liang Li, Chih-kuan Yeh, Hootan Nakhost, Yasuhisa Fujii,
  Alex Ratner, Ranjay Krishna, Chen-Yu Lee, and Tomas Pfister. 2023.
\newblock \href {https://doi.org/10.18653/v1/2023.findings-acl.507} {Distilling
  step-by-step! outperforming larger language models with less training data
  and smaller model sizes}.
\newblock In \emph{Findings of the Association for Computational Linguistics:
  ACL 2023}, pages 8003--8017, Toronto, Canada. Association for Computational
  Linguistics.

\bibitem[{Huang et~al.(2023)Huang, Dong, Wang, Hao, Singhal, Ma, Lv, Cui,
  Mohammed, Liu et~al.}]{visual6B}
Shaohan Huang, Li~Dong, Wenhui Wang, Yaru Hao, Saksham Singhal, Shuming Ma,
  Tengchao Lv, Lei Cui, Owais~Khan Mohammed, Qiang Liu, et~al. 2023.
\newblock Language is not all you need: Aligning perception with language
  models.
\newblock \emph{arXiv preprint arXiv:2302.14045}.

\bibitem[{Huang et~al.(2022)Huang, Chen, Yu, and McKeown}]{ildis}
Yukun Huang, Yanda Chen, Zhou Yu, and Kathleen McKeown. 2022.
\newblock In-context learning distillation: Transferring few-shot learning
  ability of pre-trained language models.
\newblock \emph{arXiv preprint arXiv:2212.10670}.

\bibitem[{Jiang et~al.(2023)Jiang, Chan, Chen, and Wang}]{lion}
Yuxin Jiang, Chunkit Chan, Mingyang Chen, and Wei Wang. 2023.
\newblock Lion: Adversarial distillation of closed-source large language model.
\newblock \emph{arXiv preprint arXiv:2305.12870}.

\bibitem[{Lefaudeux et~al.(2022)Lefaudeux, Massa, Liskovich, Xiong, Caggiano,
  Naren, Xu, Hu, Tintore, Zhang, Labatut, and Haziza}]{xformer}
Benjamin Lefaudeux, Francisco Massa, Diana Liskovich, Wenhan Xiong, Vittorio
  Caggiano, Sean Naren, Min Xu, Jieru Hu, Marta Tintore, Susan Zhang, Patrick
  Labatut, and Daniel Haziza. 2022.
\newblock xformers: A modular and hackable transformer modelling library.
\newblock \url{https://github.com/facebookresearch/xformers}.

\bibitem[{Li et~al.(2023{\natexlab{a}})Li, Zhang, Chen, Wang, Yang, and
  Liu}]{otter}
Bo~Li, Yuanhan Zhang, Liangyu Chen, Jinghao Wang, Jingkang Yang, and Ziwei Liu.
  2023{\natexlab{a}}.
\newblock Otter: A multi-modal model with in-context instruction tuning.
\newblock \emph{arXiv preprint arXiv:2305.03726}.

\bibitem[{Li et~al.(2023{\natexlab{b}})Li, Wang, Wang, Ge, Ge, and
  Shan}]{seedbench}
Bohao Li, Rui Wang, Guangzhi Wang, Yuying Ge, Yixiao Ge, and Ying Shan.
  2023{\natexlab{b}}.
\newblock Seed-bench: Benchmarking multimodal llms with generative
  comprehension.
\newblock \emph{arXiv preprint arXiv:2307.16125}.

\bibitem[{Li et~al.(2023{\natexlab{c}})Li, Li, Savarese, and Hoi}]{blip2}
Junnan Li, Dongxu Li, Silvio Savarese, and Steven Hoi. 2023{\natexlab{c}}.
\newblock Blip-2: Bootstrapping language-image pre-training with frozen image
  encoders and large language models.
\newblock \emph{arXiv preprint arXiv:2301.12597}.

\bibitem[{Li et~al.(2022)Li, Chen, Shen, Chen, Zhang, Li, Wang, Qian, Peng, Mao
  et~al.}]{mtcot}
Shiyang Li, Jianshu Chen, Yelong Shen, Zhiyu Chen, Xinlu Zhang, Zekun Li, Hong
  Wang, Jing Qian, Baolin Peng, Yi~Mao, et~al. 2022.
\newblock Explanations from large language models make small reasoners better.
\newblock \emph{arXiv preprint arXiv:2210.06726}.

\bibitem[{Liu et~al.(2023)Liu, Li, Wu, and Lee}]{llava}
Haotian Liu, Chunyuan Li, Qingyang Wu, and Yong~Jae Lee. 2023.
\newblock Visual instruction tuning.
\newblock \emph{arXiv preprint arXiv:2304.08485}.

\bibitem[{Lu et~al.(2023)Lu, Zhang, Wu, Gao, Gan, Zhang, Song, and
  Zhang}]{ziyavisual}
Junyu Lu, Dixiang Zhang, Xiaojun Wu, Xinyu Gao, Ruyi Gan, Jiaxing Zhang, Yan
  Song, and Pingjian Zhang. 2023.
\newblock Ziya-vl: Bilingual large vision-language model via multi-task
  instruction tuning.
\newblock \emph{arXiv preprint arXiv:2310.08166}.

\bibitem[{Lu et~al.(2022)Lu, Mishra, Xia, Qiu, Chang, Zhu, Tafjord, Clark, and
  Kalyan}]{scienceqa}
Pan Lu, Swaroop Mishra, Tanglin Xia, Liang Qiu, Kai-Wei Chang, Song-Chun Zhu,
  Oyvind Tafjord, Peter Clark, and Ashwin Kalyan. 2022.
\newblock Learn to explain: Multimodal reasoning via thought chains for science
  question answering.
\newblock \emph{Advances in Neural Information Processing Systems},
  35:2507--2521.

\bibitem[{OpenAI(2023{\natexlab{a}})}]{chatgpt}
OpenAI. 2023{\natexlab{a}}.
\newblock Chatgpt.
\newblock https://chat.openai.com.

\bibitem[{OpenAI(2023{\natexlab{b}})}]{gpt4}
OpenAI. 2023{\natexlab{b}}.
\newblock Gpt-4 technical report.
\newblock \emph{arXiv preprint arXiv:2303.08774}.

\bibitem[{Ouyang et~al.(2022{\natexlab{a}})Ouyang, Wu, Jiang, Almeida,
  Wainwright, Mishkin, Zhang, Agarwal, Slama, Ray et~al.}]{instructgpt}
Long Ouyang, Jeffrey Wu, Xu~Jiang, Diogo Almeida, Carroll Wainwright, Pamela
  Mishkin, Chong Zhang, Sandhini Agarwal, Katarina Slama, Alex Ray, et~al.
  2022{\natexlab{a}}.
\newblock Training language models to follow instructions with human feedback.
\newblock \emph{Advances in Neural Information Processing Systems},
  35:27730--27744.

\bibitem[{Ouyang et~al.(2022{\natexlab{b}})Ouyang, Wu, Jiang, Almeida,
  Wainwright, Mishkin, Zhang, Agarwal, Slama, Ray et~al.}]{trainhuman}
Long Ouyang, Jeffrey Wu, Xu~Jiang, Diogo Almeida, Carroll Wainwright, Pamela
  Mishkin, Chong Zhang, Sandhini Agarwal, Katarina Slama, Alex Ray, et~al.
  2022{\natexlab{b}}.
\newblock Training language models to follow instructions with human feedback.
\newblock \emph{Advances in Neural Information Processing Systems(NIPS)},
  35:27730--27744.

\bibitem[{Radford et~al.(2021)Radford, Kim, Hallacy, Ramesh, Goh, Agarwal,
  Sastry, Askell, Mishkin, Clark et~al.}]{clip}
Alec Radford, Jong~Wook Kim, Chris Hallacy, Aditya Ramesh, Gabriel Goh,
  Sandhini Agarwal, Girish Sastry, Amanda Askell, Pamela Mishkin, Jack Clark,
  et~al. 2021.
\newblock Learning transferable visual models from natural language
  supervision.
\newblock In \emph{International conference on machine learning}, pages
  8748--8763. PMLR.

\bibitem[{Radford et~al.(2019)Radford, Wu, Child, Luan, Amodei, Sutskever
  et~al.}]{radford2019language}
Alec Radford, Jeffrey Wu, Rewon Child, David Luan, Dario Amodei, Ilya
  Sutskever, et~al. 2019.
\newblock Language models are unsupervised multitask learners.
\newblock \emph{OpenAI blog}, 1(8):9.

\bibitem[{Raffel et~al.(2020)Raffel, Shazeer, Roberts, Lee, Narang, Matena,
  Zhou, Li, and Liu}]{raffel2020exploring}
Colin Raffel, Noam Shazeer, Adam Roberts, Katherine Lee, Sharan Narang, Michael
  Matena, Yanqi Zhou, Wei Li, and Peter~J Liu. 2020.
\newblock Exploring the limits of transfer learning with a unified text-to-text
  transformer.
\newblock \emph{The Journal of Machine Learning Research (J. Mach. Learn.
  Res)}, 21(1):5485--5551.

\bibitem[{Schuhmann et~al.(2021)Schuhmann, Vencu, Beaumont, Kaczmarczyk,
  Mullis, Katta, Coombes, Jitsev, and Komatsuzaki}]{laion}
Christoph Schuhmann, Richard Vencu, Romain Beaumont, Robert Kaczmarczyk,
  Clayton Mullis, Aarush Katta, Theo Coombes, Jenia Jitsev, and Aran
  Komatsuzaki. 2021.
\newblock Laion-400m: Open dataset of clip-filtered 400 million image-text
  pairs.
\newblock \emph{arXiv preprint arXiv:2111.02114}.

\bibitem[{Shi et~al.(2022)Shi, Suzgun, Freitag, Wang, Srivats, Vosoughi, Chung,
  Tay, Ruder, Zhou et~al.}]{cotreasoners}
Freda Shi, Mirac Suzgun, Markus Freitag, Xuezhi Wang, Suraj Srivats, Soroush
  Vosoughi, Hyung~Won Chung, Yi~Tay, Sebastian Ruder, Denny Zhou, et~al. 2022.
\newblock Language models are multilingual chain-of-thought reasoners.
\newblock \emph{arXiv preprint arXiv:2210.03057}.

\bibitem[{Touvron et~al.(2023)Touvron, Lavril, Izacard, Martinet, Lachaux,
  Lacroix, Rozi{\`e}re, Goyal, Hambro, Azhar et~al.}]{llama}
Hugo Touvron, Thibaut Lavril, Gautier Izacard, Xavier Martinet, Marie-Anne
  Lachaux, Timoth{\'e}e Lacroix, Baptiste Rozi{\`e}re, Naman Goyal, Eric
  Hambro, Faisal Azhar, et~al. 2023.
\newblock Llama: Open and efficient foundation language models.
\newblock \emph{arXiv preprint arXiv:2302.13971}.

\bibitem[{Vicente et~al.(2016)Vicente, Hou, Yu, Hoai, and Samaras}]{sbu}
Tom{\'a}s F~Yago Vicente, Le~Hou, Chen-Ping Yu, Minh Hoai, and Dimitris
  Samaras. 2016.
\newblock Large-scale training of shadow detectors with noisily-annotated
  shadow examples.
\newblock In \emph{Computer Vision--ECCV 2016: 14th European Conference,
  Amsterdam, The Netherlands, October 11-14, 2016, Proceedings, Part VI 14},
  pages 816--832. Springer.

\bibitem[{Wang et~al.(2023{\natexlab{a}})Wang, Ge, Ding, Kankanhalli, and
  Shan}]{GVT}
Guangzhi Wang, Yixiao Ge, Xiaohan Ding, Mohan Kankanhalli, and Ying Shan.
  2023{\natexlab{a}}.
\newblock What makes for good visual tokenizers for large language models?
\newblock \emph{arXiv preprint arXiv:2305.12223}.

\bibitem[{Wang et~al.(2023{\natexlab{b}})Wang, Li, Chen, Zhu, Lin, Cao, Liu,
  Liu, and Sui}]{llmeva}
Peiyi Wang, Lei Li, Liang Chen, Dawei Zhu, Binghuai Lin, Yunbo Cao, Qi~Liu,
  Tianyu Liu, and Zhifang Sui. 2023{\natexlab{b}}.
\newblock Large language models are not fair evaluators.
\newblock \emph{arXiv preprint arXiv:2305.17926}.

\bibitem[{Wang et~al.(2023{\natexlab{c}})Wang, Zhu, and Wang}]{incontext2}
Xinyi Wang, Wanrong Zhu, and William~Yang Wang. 2023{\natexlab{c}}.
\newblock Large language models are implicitly topic models: Explaining and
  finding good demonstrations for in-context learning.
\newblock \emph{arXiv preprint arXiv:2301.11916}.

\bibitem[{Wang et~al.(2022{\natexlab{a}})Wang, Wei, Schuurmans, Le, Chi,
  Narang, Chowdhery, and Zhou}]{selfsonsistency}
Xuezhi Wang, Jason Wei, Dale Schuurmans, Quoc~V Le, Ed~H Chi, Sharan Narang,
  Aakanksha Chowdhery, and Denny Zhou. 2022{\natexlab{a}}.
\newblock Self-consistency improves chain of thought reasoning in language
  models.
\newblock In \emph{The Eleventh International Conference on Learning
  Representations}.

\bibitem[{Wang et~al.(2022{\natexlab{b}})Wang, Kordi, Mishra, Liu, Smith,
  Khashabi, and Hajishirzi}]{selfinstruct}
Yizhong Wang, Yeganeh Kordi, Swaroop Mishra, Alisa Liu, Noah~A Smith, Daniel
  Khashabi, and Hannaneh Hajishirzi. 2022{\natexlab{b}}.
\newblock Self-instruct: Aligning language model with self generated
  instructions.
\newblock \emph{arXiv preprint arXiv:2212.10560}.

\bibitem[{Wei et~al.(2022)Wei, Wang, Schuurmans, Bosma, Xia, Chi, Le, Zhou
  et~al.}]{cotprompt}
Jason Wei, Xuezhi Wang, Dale Schuurmans, Maarten Bosma, Fei Xia, Ed~Chi, Quoc~V
  Le, Denny Zhou, et~al. 2022.
\newblock Chain-of-thought prompting elicits reasoning in large language
  models.
\newblock \emph{Advances in Neural Information Processing Systems(NIPS)},
  35:24824--24837.

\bibitem[{Wu et~al.(2023)Wu, Waheed, Zhang, Abdul-Mageed, and Aji}]{lamamini}
Minghao Wu, Abdul Waheed, Chiyu Zhang, Muhammad Abdul-Mageed, and Alham~Fikri
  Aji. 2023.
\newblock Lamini-lm: A diverse herd of distilled models from large-scale
  instructions.
\newblock \emph{arXiv preprint arXiv:2304.14402}.

\bibitem[{Yao et~al.(2023)Yao, Aminabadi, Ruwase, Rajbhandari, Wu, Awan,
  Rasley, Zhang, Li, Holmes, Zhou, Wyatt, Smith, Kurilenko, Qin, Tanaka, Che,
  Song, and He}]{deepspeed}
Zhewei Yao, Reza~Yazdani Aminabadi, Olatunji Ruwase, Samyam Rajbhandari,
  Xiaoxia Wu, Ammar~Ahmad Awan, Jeff Rasley, Minjia Zhang, Conglong Li, Connor
  Holmes, Zhongzhu Zhou, Michael Wyatt, Molly Smith, Lev Kurilenko, Heyang Qin,
  Masahiro Tanaka, Shuai Che, Shuaiwen~Leon Song, and Yuxiong He. 2023.
\newblock {DeepSpeed-Chat: Easy, Fast and Affordable RLHF Training of
  ChatGPT-like Models at All Scales}.
\newblock \emph{arXiv preprint arXiv:2308.01320}.

\bibitem[{Ye et~al.(2023)Ye, Xu, Xu, Ye, Yan, Zhou, Wang, Hu, Shi, Shi
  et~al.}]{mplug}
Qinghao Ye, Haiyang Xu, Guohai Xu, Jiabo Ye, Ming Yan, Yiyang Zhou, Junyang
  Wang, Anwen Hu, Pengcheng Shi, Yaya Shi, et~al. 2023.
\newblock mplug-owl: Modularization empowers large language models with
  multimodality.
\newblock \emph{arXiv preprint arXiv:2304.14178}.

\bibitem[{Zhang et~al.(2023{\natexlab{a}})Zhang, Han, Zhou, Hu, Yan, Lu, Li,
  Gao, and Qiao}]{llamaadapter}
Renrui Zhang, Jiaming Han, Aojun Zhou, Xiangfei Hu, Shilin Yan, Pan Lu,
  Hongsheng Li, Peng Gao, and Yu~Qiao. 2023{\natexlab{a}}.
\newblock Llama-adapter: Efficient fine-tuning of language models with
  zero-init attention.
\newblock \emph{arXiv preprint arXiv:2303.16199}.

\bibitem[{Zhang et~al.(2022)Zhang, Roller, Goyal, Artetxe, Chen, Chen, Dewan,
  Diab, Li, Lin et~al.}]{zhang2022opt}
Susan Zhang, Stephen Roller, Naman Goyal, Mikel Artetxe, Moya Chen, Shuohui
  Chen, Christopher Dewan, Mona Diab, Xian Li, Xi~Victoria Lin, et~al. 2022.
\newblock Opt: Open pre-trained transformer language models.
\newblock \emph{arXiv preprint arXiv:2205.01068}.

\bibitem[{Zhang et~al.(2023{\natexlab{b}})Zhang, Zhang, Li, Zhao, Karypis, and
  Smola}]{mmcot}
Zhuosheng Zhang, Aston Zhang, Mu~Li, Hai Zhao, George Karypis, and Alex Smola.
  2023{\natexlab{b}}.
\newblock Multimodal chain-of-thought reasoning in language models.
\newblock \emph{arXiv preprint arXiv:2302.00923}.

\bibitem[{Zhu et~al.(2023{\natexlab{a}})Zhu, Chen, Shen, Li, and
  Elhoseiny}]{minigpt4}
Deyao Zhu, Jun Chen, Xiaoqian Shen, Xiang Li, and Mohamed Elhoseiny.
  2023{\natexlab{a}}.
\newblock Minigpt-4: Enhancing vision-language understanding with advanced
  large language models.
\newblock \emph{arXiv preprint arXiv:2304.10592}.

\bibitem[{Zhu et~al.(2023{\natexlab{b}})Zhu, Li, Liu, Ma, and Wang}]{dismodel}
Xunyu Zhu, Jian Li, Yong Liu, Can Ma, and Weiping Wang. 2023{\natexlab{b}}.
\newblock A survey on model compression for large language models.
\newblock \emph{arXiv preprint arXiv:2308.07633}.

\end{thebibliography}
\bibliographystyle{acl_natbib}




\end{document}